\documentclass{article}

\usepackage{enumitem}
\usepackage[utf8]{inputenc} 
\usepackage[T1]{fontenc}    
\usepackage{hyperref}       

\usepackage{url}            
\usepackage{booktabs}       
\usepackage{amsfonts}       
\usepackage{nicefrac}       
\usepackage{microtype}      
\usepackage{graphicx}
\usepackage{amsmath,amsfonts,amssymb,amsthm}
\usepackage{color}
\usepackage{subcaption}
\usepackage{amssymb} 
\usepackage{pifont} 
\usepackage{lipsum}
\usepackage{natbib}
\usepackage{xr}
\usepackage{wrapfig}
\usepackage{multirow}
\usepackage{tabu}
\usepackage{graphicx}
\usepackage{tabularx}
\usepackage{textcomp}
\usepackage{xcolor}
\usepackage{booktabs} 
\usepackage{subcaption}
\usepackage{makecell}
\usepackage{colortbl}

\usepackage[accepted]{icml2021}

\icmltitlerunning{\methodname: Enhancing saliency maps using decoys}

\newcommand{\ie}{\mbox{\it{i.e.,\ }}}
\newcommand{\eg}{\mbox{\it{e.g.,\ }}}
\newcommand{\quotes}[1]{``#1''}

\newcommand{\dperp}{\mathbin{\rotatebox[origin=c]{90}{$\models$}}}

\newcommand{\methodname}{DANCE} 

\newcommand{\bx}{\mathbf{x}}

\newcommand{\bW}{\mathbf{W}}
\newcommand{\bb}{\mathbf{b}}
\newcommand{\bo}{\mathbf{o}}
\newcommand{\bp}{\mathbf{p}}

\newcommand{\bg}{\mathbf{g}}
\newcommand{\bM}{\mathbf{m}}

\newcommand{\bH}{\mathbf{H}}

\theoremstyle{proposition}
\newtheorem{prop}{Proposition}

\theoremstyle{theorem}
\newtheorem{thm}{Theorem}

\begin{document}

\twocolumn[
\icmltitle{\methodname{}: Enhancing saliency maps using decoys }



\icmlsetsymbol{equal}{*}

\begin{icmlauthorlist}
\icmlauthor{Yang Young Lu}{equal,uw}
\icmlauthor{Wenbo Guo}{equal,psu}
\icmlauthor{Xinyu Xing}{psu}
\icmlauthor{William Stafford Noble}{uw2}
\end{icmlauthorlist}

\icmlaffiliation{uw}{Department of Genome Sciences, University of Washington, Seattle, WA, USA}
\icmlaffiliation{psu}{College of Information Sciences and Technology, The Pennsylvania State University, State College, PA, USA}
\icmlaffiliation{uw2}{Paul G. Allen School of Computer Science and Engineering, University of Washington, Seattle, WA, USA}

\icmlcorrespondingauthor{Xinyu Xing}{xxing@ist.psu.edu}
\icmlcorrespondingauthor{William Stafford Noble}{william-noble@uw.edu}

\icmlkeywords{Machine Learning, ICML}

\vskip 0.3in
]



\printAffiliationsAndNotice{\icmlEqualContribution} 

\vspace{20pt}

\begin{abstract}
Saliency methods can make deep neural network predictions more interpretable by identifying a set of critical features in an input sample, such as pixels that contribute most strongly to a prediction made by an image classifier. Unfortunately, recent evidence suggests that many saliency methods poorly perform, especially in situations where gradients are saturated, inputs contain adversarial perturbations, or predictions rely upon inter-feature dependence. To address these issues, we propose a framework, \methodname{}, which improves the robustness of saliency methods by following a two-step procedure. First, we introduce a perturbation mechanism that subtly varies the input sample without changing its intermediate representations. Using this approach, we can gather a corpus of perturbed (``decoy'') data samples while ensuring that the perturbed and original input samples follow similar distributions. Second, we compute saliency maps for the decoy samples and propose a new method to aggregate saliency maps. With this design, we offset influence of gradient saturation. From a theoretical perspective, we show that the aggregated saliency map not only captures inter-feature dependence but, more importantly, is robust against previously described adversarial perturbation methods. Our empirical results suggest that, both qualitatively and quantitatively, \methodname{} outperforms existing methods in a variety of application domains.
\footnote{The Apache licensed source code of \methodname{} will be available at 
\url{https://bitbucket.org/noblelab/dance}. }
\end{abstract}

\section{Introduction}
\label{sec:introduction}
Deep neural networks (DNNs) deliver remarkable performance in an increasingly wide range of application domains, but they often do so in an inscrutable fashion, delivering predictions without accompanying explanations. In a practical setting such as automated analysis of pathology images, if a patient sample is classified as malignant, then the physician will want to know which parts of the image contribute to this diagnosis. Thus, in general, a DNN that delivers interpretations alongside its predictions will enhance the credibility and utility of its predictions for end users \citep{lipton2016mythos}.

In this paper, we focus on a popular branch of explanation methods, often referred to as {\em saliency methods}, which aim to find input features (\eg image pixels or words) that strongly influence the network predictions \citep{simonyan2013deep, selvaraju2016grad, binder2016layer, shrikumar2017learning, smilkov2017smoothgrad, sundararajan2017axiomatic, ancona2017towards}. Saliency methods typically rely on back-propagation from the network's output back to its input to assign a saliency score to individual features so that higher scores indicate higher importance to the output prediction. Despite attracting increasing attention, saliency methods suffer from several fundamental limitations:
\begin{itemize}[leftmargin=1em]
  \item \textbf{Gradient saturation} \citep{sundararajan2017axiomatic, shrikumar2017learning, smilkov2017smoothgrad} may lead to the problem that the gradients of important features have small magnitudes, breaking down the implicit assumption that important features, in general, correspond to large gradients. This issue can be triggered when the DNN outputs are flattened in the vicinity of important features.
  \item \textbf{Importance isolation} \citep{singla2019understanding} refers to the problem that gradient-based saliency methods evaluate the feature importance in an isolated fashion, implicitly assuming that the other features are fixed.
  \item \textbf{Perturbation sensitivity} \citep{ghorbani2017interpretation,kindermans2017reliability,levine2019certifiably} refers to the observation that even imperceivable, random perturbations or a simple shift transformation of the input data may lead to a large change in the resulting saliency scores.
\end{itemize}

In this paper, we propose a novel saliency method, Decoy-enhANCEd saliency (\methodname{}), to tackle these limitations. At a high level, \methodname{} generates the saliency score of an input by aggregating the saliency scores of multiple perturbed copies of this input. Specifically, given an input sample of interest, \methodname{} first generates a population of perturbed samples, referred to as {\em decoys}, that perfectly mimic the neural network's intermediate representation of the original input. These decoys are used to model the variation of an input sample originating from either sensor noise or adversarial attacks.
The decoy construction procedure draws inspiration from the {\em knockoffs}, proposed recently by \citet{barber2015controlling} in the setting of error-controlled feature selection, where the core idea is to generate knockoff features that perfectly mimic the empirical dependence structure among the original features.

In brief, the current paper makes three primary contributions.
First, we propose a framework to perturb input samples to produce corresponding decoys that preserve the input distribution, in the sense that the intermediate representations of the original input data and the decoys are indistinguishable. We formulate decoy generation as an optimization problem, applicable to diverse deep neural network architectures.
Second, we develop a decoy-enhanced saliency score by aggregating the saliency maps of generated decoys. By design, this score naturally offsets the impact of gradient saturation. From a theoretical perspective, we show how the proposed score can simultaneously reflect the joint effects of other dependent features and achieve robustness to adversarial perturbations.
Third, we demonstrate empirically that \methodname{} outperforms existing saliency methods, both qualitatively and quantitatively, on three real-world applications. We also quantify \methodname{}'s advantage over existing saliency methods in terms of robustness against various adversarial attacks.

\section{Related work}
\label{sec:rw}
A variety of saliency methods have been proposed in the literature. Some, such as edge detectors and Guided Backpropagation \citep{springenberg2014striving} are independent of the predictive model \citep{nie2018theoretical,adebayo2018sanity}.\footnote{\citet{sixt2019explanations} shows that LRP \citep{binder2016layer} is independent of the parameters of certain layers.} Others are designed only for specific architectures (\ie Grad-CAM \citep{selvaraju2016grad} for CNNs, DeConvNet for CNNs with ReLU activations \citep{zeiler2014visualizing}). In this paper, instead of exhaustively evaluating all saliency methods, we apply our method to three saliency methods that do depend on the predictor (\ie passing the sanity checks in \citet{adebayo2018sanity} and \citet{sixt2019explanations}) and are applicable to diverse DNN architectures.
\begin{itemize}[leftmargin=1em]
  \item The \textbf{vanilla gradient} method \citep{simonyan2013deep} calculates the gradient of the class score with respect to the input $\bx$, defined as
   \begin{equation}
   \nonumber
   E_{grad}(\bx ; F^c)=\triangledown_{\bx}F^c(\bx)
   \end{equation}
  \item \textbf{SmoothGrad} \citep{smilkov2017smoothgrad} seeks to reduce noise in the saliency map by averaging over explanations of the noisy copies of an input, defined as
   \begin{equation}
   \nonumber
   E_{sg}(\bx ; F^c)=\frac{1}{N}\sum_{i=1}^{N}E_{grad}(\bx+g_i ; F^c)
   \end{equation}
   where $g_i\sim N(0,\sigma^2)$ indicates the noise vectors.
  \item The \textbf{integrated gradient} method \citep{sundararajan2017axiomatic} starts from a baseline input $\bx^0$ and sums over the gradient with respect to scaled versions of the input ranging from the baseline to the observed input, defined as
  \begin{equation}
   \nonumber
   E_{ig}(\bx ; F^c)=(\bx-\bx^0)\times \int_{0}^{1} \triangledown_{\bx}F^c(\bx^0+\alpha (\bx-\bx^0)) d \alpha
   \end{equation}
  Note that $input \odot gradient$ and DeepLIFT \citep{shrikumar2017learning} are strongly related to the integrated gradient method, as shown by \citet{ancona2017towards}.
\end{itemize}

We do not empirically compare to several other categories of methods.
{\em Counterfactual-based methods} work under the same setup as saliency methods, providing explanations for the predictions of a pre-trained DNN model \citep{sturmfels2020visualizing}. These methods identify the important subregions within an input image by perturbing the subregions (by adding noise, rescaling \citep{sundararajan2017axiomatic}, blurring \citep{fong2017interpretable}, or inpainting \citep{change2019xplaining}) and measuring the resulting changes in the predictions \citep{ribeiro2016should,lundberg2017unified,chen2018learning, fong2017interpretable,dabkowski2017real,change2019xplaining,yousefzadeh2019interpreting,goyal2019counterfactual}. Although these methods do identify meaningful subregions in practice, they exhibit several limitations. First, counterfactual-based methods implicitly assume that regions containing the object most contribute to the prediction \citep{fan2017adversarial}. However, \citet{moosavi2017universal} showed that counterfactual-based methods are also vulnerable to adversarial attacks, which force these methods to output unrelated background rather than the meaningful objects as important subregions. Second, the counterfactual images may be potentially far away from the training distribution, causing ill-defined classifier behavior \citep{burns2019hypothesis,hendrycks2019benchmarking}.
\begin{figure*}[ht]
\centering
\includegraphics[width=1.0\textwidth]{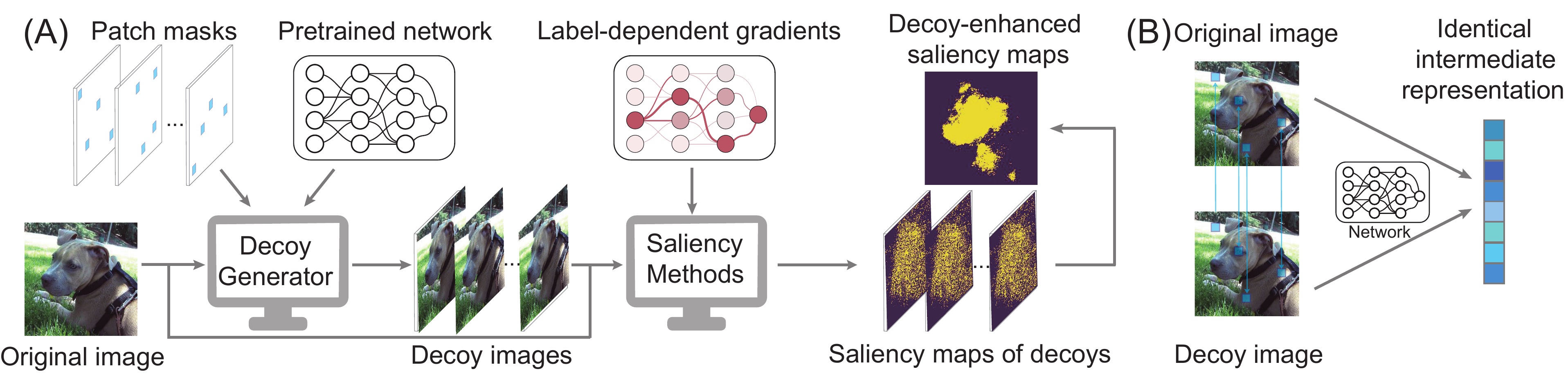}
\caption{{\bf Overview of \methodname{}.} (A) The \methodname{} workflow. (B) The swapping operation between original and decoy images.}
\label{fig:overflow}
\end{figure*}

In addition to these limitations, counterfactual-based methods and our decoy-based method are fundamentally different in three ways. First, the former seeks the minimum set of features to exclude in order to minimize the prediction score or to include in order to maximize the prediction score \citep{fong2017interpretable}, whereas our approach aims to characterize the influence of each feature on the prediction score. Second, counterfactual-based methods explicitly consider the decision boundary by comparing each image to the closest image on the other side of the boundary. In contrast, the proposed method only considers the decision boundary implicitly by calculating the gradient's variants. Third, unlike counterfactual images, which could potentially be out-of-distribution, decoys are plausibly constructed in the sense that their intermediate representations are indistinguishable from the original input data by design. Because of these limitations and differences, we do not compare our method with counterfactual-based methods.

In addition to saliency methods and counterfactual-based methods, several other types of interpretation methods have been proposed that either aim for a different goal or have a different setup. For example, recent research (\eg \citet{ribeiro2016should, lundberg2017unified, chen2018learning, chen2018shapley}) designed techniques to explain a black-box model, where the model's internal weights are inaccessible. \citet{koh2017understanding} and some follow-up work \citep{yeh2018representer, koh2019accuracy} tried to find the training points that are most influential for a given test sample. Some other efforts have been made to train a more interpretable DNN classifier \citep{fan2017adversarial, zolna2019classifier, alvarez2018towards, toneva2019interpreting}, synthesize samples that represent the model predictions \citep{ghorbani2019towards,chen2019looks}, or identifying noise-tolerant features \citep{ikeno2018maximizing,schulz2020restricting}. However, due to the task and setup differences, we do not consider these methods in this paper.

\section{Methods}

\subsection{Problem setup}
\label{sec:background}
Consider a multi-label classification task in which a pre-trained neural network model implements a function $F$: $\mathbb{R}^d \mapsto \mathbb{R}^C$ that maps from the given input $\bx \in \mathbb{R}^d$ to $C$ predicted classes. The score for each class $c \in \left \{ 1,\cdots,C \right \}$ is $F^c(\bx)$, and the predicted class is the one with maximum score, \ie $\arg \max_{c \in \left \{ 1,\cdots,C \right \}} F^c(\bx)$. A {\em saliency method} aims to assign to each feature a {\em saliency score}, encoded in a saliency map $E(\bx ; F^c): \mathbb{R}^d \mapsto \mathbb{R}^d$, in which the features with higher scores represent higher \quotes{importance} relative to the final prediction.

Given a pre-trained neural network model $F$ with $L$ layers, an input $\bx$, and a saliency method $E$ such that $E(\bx ; F)$ is a saliency map of the same dimensions as $\bx$, \methodname{} operates in two steps: generating decoys and aggregating the saliency maps of the decoys (Figure~\ref{fig:overflow}A).

\subsection{Decoy definition}
\label{sec:decoydef}
Say that $F_\ell: \mathbb{R}^d \mapsto \mathbb{R}^{d_\ell}$ is the function instantiated by the given network, which maps from an input $\bx \in \mathbb{R}^d$ to its intermediate representation $F_\ell(\bx) \in \mathbb{R}^{d_\ell}$ at layer $\ell \in \left \{ 1,2,\cdots,L \right \}$. A vector $\tilde{\bx} \in \mathbb{R}^d$ is said to be a {\em decoy} of $\bx \in \mathbb{R}^d$ at a specified layer $\ell$ if the following swappable condition is satisfied:
\begin{equation}
\begin{aligned}
& F_\ell(\bx)=F_\ell(\bx_{\text{swap}(\tilde{\bx}, \mathcal{K})}), \; \\
& \text{for swappable features }\; \mathcal{K} \subset \{1,\cdots, d\} \, .
\end{aligned}
\label{eq:swappable}
\end{equation}
Here, the $\text{swap}(\tilde{\bx},\mathcal{K})$ operation swaps features between $\bx$ and $\tilde{\bx}$ based on the elements in $\mathcal{K}$. In this work, $\mathcal{K}$ represents a small meaningful feature set, which represents a small region/segment in an image or a group of words (embeddings) in a sentence. Take an image recognition task for example. Assume $\mathcal{K}= \{10\}$ and $\tilde{\bx}$ is a zero matrix, then $\bx_{\text{swap}(\tilde{\bx}, \mathcal{K})}$ indicates a new image that is identical to $\bx$ except that the tenth pixel is set to zero. An illustrative explanation of a swap operator is shown in Figure~\ref{fig:overflow}(B).

Using the swappable condition, we aim to ensure that the original image $\bx$ and its decoy $\tilde{\bx}$ are indistinguishable in terms of the intermediate representation at layer $\ell$. Note in particular that the construction of decoys relies solely on the first $\ell$ layers of the neural network $F_1,F_2,\cdots,F_\ell$ and is independent of the succeeding layers $F_{\ell+1},\cdots,F_{L}$. As such, $\tilde{\bx}$ is conditionally independent of the classification task $F(\bx)$ given the input $\bx$; \ie $\tilde{\bx} \dperp F(\bx)|\bx$.

\subsection{Decoy generation}
\label{sec:decoygen}
To identify decoys satisfying the swappable condition, we solve the following optimization problem:
\begin{equation}
\label{eq:decoygen1}
\begin{aligned}
\text{maximize}_{\tilde{\bx} \in [\bx_{\text{min}}, \bx_{\text{max}}]^{d}}  & \: \: \left \| ((\tilde{\bx}-\bx)\cdot s)^+ \right \|_1, \\
\text{s.t.} & \: \: \left\{\begin{matrix}
\left \| F_\ell(\tilde{\bx}) - F_\ell(\bx) \right \|_\infty \leq \epsilon, \\
(\tilde{\bx} - \bx) \circ (1-\mathcal{M})=0
\end{matrix}\right.
\end{aligned}
\end{equation}
Here, $(\cdot)^+=\max(\cdot,0)$, and the operators $\left \| \cdot  \right \|_1$ and $\left \| \cdot  \right \|_\infty$ correspond to the $L_1$ and $L_\infty$ norms, respectively. $\mathcal{M} \in \left \{ 0,1 \right \}^{d}$ is a specified binary mask, where $\mathcal{M}_i=0$ indicates that the $i$th features of $\bx$ and $\tilde{\bx}$ are kept the same (realized by the constraint $(\tilde{\bx} - \bx) \circ (1-\mathcal{M})=0$). In other words, we take $\tilde{\bx}$ and $\bx$ to be indistinguishable except for the swappable features indicated by the mask (\ie $\bx_{swap(\tilde{\bx}, (1-\mathcal{M}))} = \tilde{\bx}$). The value of each feature in the decoy $\tilde{\bx}$ is restricted to lie in a legitimate value range \ie $[\bx_{\text{min}}, \bx_{\text{max}}]$ (\eg the pixel values should lie in $[0, 255]$). We further impose the constraint $\left \| F_\ell(\tilde{\bx}) - F_\ell(\bx) \right \|_\infty \leq \epsilon$, which ensures that the generated decoy satisfies the swappable condition described in Equation~\eqref{eq:swappable}.

As illustrated in Figure~\ref{fig:overflow}, a population of $n$ patch masks are constructed subject to the principle that each swappable patch is covered at least once. Because each swappable patch is small (\eg a small region/segment in an image), assigning each patch mask to a single patch would be computationally expensive. Accordingly, we aggregate multiple patches into a combined patch mask for computational efficiency (see Supplementary Section S1 for details).  Empirical results suggest that \methodname{} is robust to the number of patches that are aggregated into each mask (Figure~\ref{fig:imagenet}C).

As is shown later in Section~\ref{sec:decoysaliency}, \methodname{} aims to capture the range of the saliency maps among all decoys. To achieve this, we first need to estimate the range of values among the decoys by estimating the range of perturbation values that can be added to the input without violating the swappable condition. In other words, we maximize the deviation between $\tilde{\bx}$ and $\bx$ from both the positive and negative directions, \ie $s=+1$ and $s=-1$.
As shown in Equation~\eqref{eq:decoygen1}, for each specified mask $\mathcal{M}$, we compute two decoys---one for the positive deviation (\ie $s=+1$) and the other for the negative one (\ie $s=-1$). More details about how to optimize Equation~\ref{eq:decoygen1} can be found in Supplementary Section S1.

\subsection{Decoy-enhanced saliency scores}
\label{sec:decoysaliency}

We denote the generated decoys as $\left \{ \tilde{\bx}^1, \tilde{\bx}^2,\cdots,\tilde{\bx}^{2n} \right \}$, \ie $n$ decoys in the positive direction and $n$ in the negative direction. For these decoys, we then apply a given saliency method $E$ to yield the corresponding decoy saliency maps $\left \{ E(\tilde{\bx}^1; F), E(\tilde{\bx}^2; F),\cdots,E(\tilde{\bx}^{2n}; F) \right \}$. With these decoy saliency maps in hand, for each feature $\bx_{i}$ in $\bx$, we characterize its saliency score variation by using a population of saliency scores $\tilde{E}_i = \left \{ E(\tilde{\bx}^1; F^c)_i, E(\tilde{\bx}^2; F^c)_i,\cdots,E(\tilde{\bx}^{2n}; F^c)_i \right \}$. Here we define the decoy-enhanced saliency score $Z_i$ for each feature $\bx_i$ as
\begin{equation}
Z_i=\max ( \tilde{E}_i ) - \min ( \tilde{E}_i ) \, .
\label{eq:saliencystat}
\end{equation}
Here, $Z_i$ is determined by the empirical range of the decoy saliency scores. Ideally, important features will have large values and unimportant ones will have small values. 

\subsection{Theoretical insights}
\label{sec:theory}
In this section, we analyze the saliency score method in a theoretical fashion.  For expedience of exposition, we carry out the theoretical analysis using the vanilla gradient as the base saliency method. In particular, we take a convolutional neural network with the ReLU activation function as an example to discuss why the proposed interpretation method can account for inter-feature dependence while also improving explanatory robustness. It should be noted that, while we conduct our theoretical analysis in the setting of convolutional neural networks (CNNs) with a specific activation function, the conclusions drawn from the theoretical analysis can be extended to other feed-forward neural architectures and other activation functions (See Supplementary Section S4 for more details).

Consider a CNN with $L$ hidden blocks, with each layer $\ell$ containing a convolutional layer with a filter of size $\sqrt{s_\ell} \times \sqrt{s_\ell}$ and a max pooling layer with pooling size $\sqrt{s_\ell} \times \sqrt{s_\ell}$. (We set the pooling size the same as the kernel size in each block for simplicity.) The input to this CNN is $\bx \in \mathbb{R}^{d}$, unrolled from a $\sqrt{d} \times \sqrt{d}$ matrix. Similarly, we also unroll each convolutional filter into $\bg_{\ell} \in \mathbb{R}^{s_{\ell}}$, where $\bg_{\ell}$ is indexed as $(\bg_{\ell})_{j}$ for $j \in \mathcal{J}_{\ell}$. Here, $\mathcal{J}_{\ell}$ corresponds to the index shift in matrix form from the top-left to bottom-right element. For example, a $3 \times 3$ convolutional filter (\ie $s_{\ell} = 9$) is indexed by $\mathcal{J}_{\ell} = $ $\{ -\sqrt{d}-1,-\sqrt{d},-\sqrt{d}+1,-1,0,1,\sqrt{d}-1,\sqrt{d},\sqrt{d}+1 \}$. The output of the network is the probability vector $\bp \in \mathbb{R}^{C}$ generated by the softmax function, where $C$ is the total number of classes. Such a network can be represented as
\vspace{3pt}
\begin{equation}
\nonumber
\begin{aligned}
& \bM_{\ell} = \text{pool}(\text{relu}(\bg_{\ell} \ast \bM_{\ell-1})) \ \ \text{for} \ \ {\ell} = 1, 2, 3, ..., L \, , \\
& \bo = \bW_{L+1}^T \bM_{L}+\bb_{L+1},  \\
& \bp = \text{softmax}(\bo) \, ,
\end{aligned}
\end{equation}
\vspace{3pt}
where $\text{relu}(\cdot)$ and $\text{pool}(\cdot)$ indicate the ReLU and pooling operators, $\bM_{\ell} \in \mathbb{R}^{d_{\ell}}$ is the output of the block ${\ell}$ ($\bM_{0} = \bx$), and $(\bg_{\ell} \ast \bM_{\ell-1}) \in \mathbb{R}^{d_{\ell-1}}$ represents a convolutional operation on that block. We assume for simplicity that the convolution retains the input shape.

Consider an input $\bx$ and its decoy $\tilde{\bx}$, generated by swapping features in $\mathcal{K}$. For each feature $i \in \mathcal{K}$, we have the following theorem for the decoy-enhanced saliency score $Z_i$:
\begin{thm}
\emph{ In the aforementioned setting, $Z_i$ is bounded by}
\begin{equation}
\left | Z_{i} - \frac{1}{2}\left|  \sum_{k \in \mathcal{K}} (\tilde{\bx}_{k}^{+} - \tilde{\bx}_{k}^{-}) (\bH_{\bx})_{k, i}\right| \right| \leq C_1 \, .
\end{equation}
\label{thm:1}
\end{thm}
Here, $C_1 > 0$ is a bounded constant and $\bH_{\bx}$ is the Hessian of $F^c(\bx)$ on $\bx$ where $(\bH_{\bx})_{i,k}=\frac{\partial^2 F^c}{\partial \bx_i \partial \bx_k}$.
$\tilde{\bx}^{+}$ and $\tilde{\bx}^{-}$ refer to the decoy that maximizes and minimizes $E(\tilde{\bx}; F^c)$, respectively. Theorem~\ref{thm:1} implies that the proposed saliency score is determined by the second-order Hessian ($(\bH_{\bx})_{i, k}$) in the same swappable feature set. The score explicitly models the feature dependencies in the swappable feature set via this second-order Hessian, potentially capturing meaningful patterns such as edges, texture, etc.

In addition to enabling representation of inter-feature dependence, Theorem~\ref{thm:1} sheds light on the robustness of the proposed saliency score against adversarial attack. To illustrate the robustness improvement of our method, we introduce the following proposition.
\begin{prop}[]
\label{prop:1}
\emph{ Given an input $\bx$ and the corresponding adversarial sample $\hat{\bx}$, if both $\left|\bx_{i} - \tilde{\bx}_{i} \right | \leq C_2 \delta_{i}$ and $\left|\hat{\bx}_{i} - \tilde{\hat{\bx}}_{i} \right | \leq C_2 \delta_{i}$ can be obtain where $C_2 > 0$ is a bounded constant and $\delta_{i} =\left|E(\hat{\bx}, F)_{i} - E(\bx, F)_{i}\right |$, then the following relation can be guaranteed.
  }
\begin{equation}
\left|(Z_{\hat{\bx}})_{i} - (Z_{\bx})_{i}\right| \leq \left|E(\hat{\bx}, F)_{i} - E(\bx, F)_{i} \right |.
\end{equation}
\end{prop}
Given an adversarial sample $\hat{\bx}$ (\ie the perturbed $\bx$), we say a saliency method is not robust against $\hat{\bx}$ if the deviation of the corresponding explanation $\delta_{i} =\left|E(\hat{\bx}, F)_{i} - E(\bx, F)_{i}\right |$ (for all $i \in \{1,2,\cdots,d\}$) is large. According to the proposition above, we can easily discover that the deviation of our decoy-enhanced saliency score is always no larger than that of other saliency methods when a certain condition is satisfied. This indicates that, when the condition holds, our saliency method can guarantee a stronger resistance to the adversarial perturbation. To ensure the conditions $\left|\bx_{i} - \tilde{\bx}_{i} \right | \leq C_2 \delta_{i}$ and $\left|\hat{\bx}_{i} - \tilde{\hat{\bx}}_{i} \right | \leq C_2 \delta_{i}$, we can further introduce the corresponding condition as a constraint to Equation~\eqref{eq:decoygen1}. In the following section, without further clarification, the saliency scores used in our evaluation are all derived with this constraint imposed. The proof and in-depth analysis of Theorem~\ref{thm:1} and Proposition~\ref{prop:1} can be found in the Supplementary Section S2 and S3.
\begin{figure*}
\centering
\includegraphics[width=1.0\textwidth]{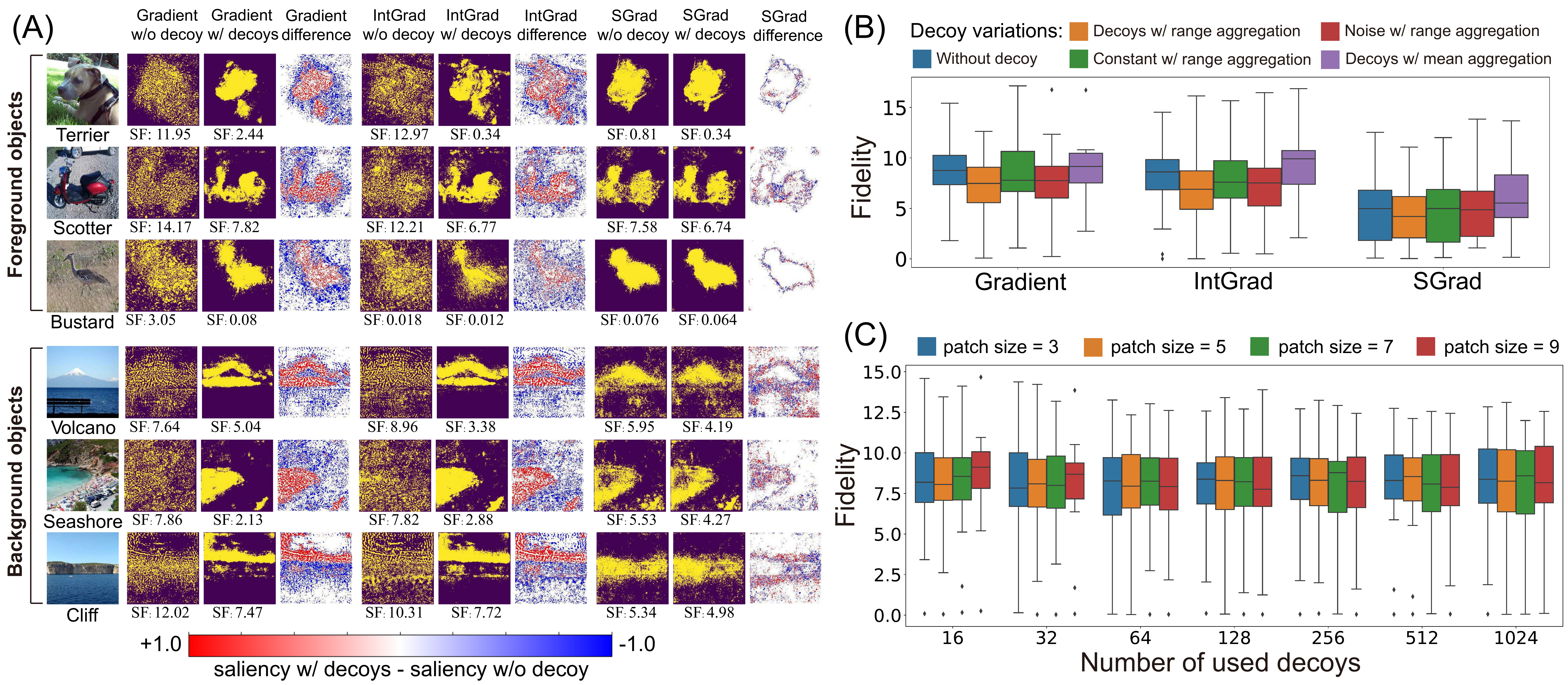}
\vspace{2pt}
\caption{{\bf Performance evaluation on ImageNet.} (A) Visualization of saliency maps on foreground and background objects. (B) Fidelity comparison of original saliency method (\ie \quotes{Without decoys}), our method (\ie \quotes{Decoys w/ range aggregation}), and its alternatives: replacing the decoy generation (Equation~\eqref{eq:decoygen1}) with constant perturbation (\ie \quotes{Constant w/ range aggregation}) or noise perturbation (\ie \quotes{Noise w/ range aggregation}); replacing the decoy aggregation (Equation~\eqref{eq:saliencystat}) with mean aggregation (\ie \quotes{Decoys w/ mean aggregation}). See Supplementary Section S14 for more statistics about the performance differences between our method and the baselines. (C) Performance with regard to variant patch size and different numbers of decoys. }
\label{fig:imagenet}
\vspace{10pt}
\end{figure*}
\section{Experiments}
\label{sec:exp}
To evaluate the effectiveness of \methodname{}, we perform extensive experiments on deep learning models that target three tasks: image classification, sentiment analysis, and network intrusion detection. Our results suggest that that \methodname{}, in conjunction with state-of-the-art saliency methods, makes already good saliency maps even more intuitively coherent.  \methodname{} also quantitatively achieves better alignment to truly important features and demonstrates stronger robustness to adversarial manipulation. The description of the datasets and experimental setup can be found in the Supplementary Section S5.

\subsection{Baseline methods }
\label{sec:exp:benchmark}

We applied \methodname{} in conjunction with three state-of-the-art saliency methods: vanilla gradient, integrated gradient, and SmoothGrad. (See Supplementary Section S9 for results from more saliency methods such as ExpGrad \citep{sturmfels2020visualizing}, VarGrad \citep{hooker2019benchmark}, and Grad-CAM \citep{selvaraju2016grad}.) As claimed in Section~\ref{sec:rw},  a prerequisite of saliency methods is the dependency on the predictor. To confirm that the conjunction with \methodname{} does not violate this prerequisite, we carried out a sanity check on the ImageNet dataset. The results (Supplementary Section S6) show that our method does indeed depend on the predictor.

One significant challenge when comparing different saliency methods is that each method produces a raw saliency map with its own distribution. Therefore, to facilitate a fair comparison among all methods, we used a consistent post-processing scheme to normalize all methods. Specifically, we selected the top-$K$ normalization, \ie constructing a binary saliency map by retaining only the top-$K$ features ranked by each method. We then set the saliency value of the selected features equal to $1$ and the remaining features equal to $0$. Here we chose $K$ as the top 20\% of all features, and we show that our results are robust to variation in the choice of $K$ in Supplementary Section S11. It is worth mentioning that in this paper we do not consider another common normalization scheme, 0-1 normalization (\ie linearly rescaling the saliency values to the range $[0, 1]$), because 0-1 normalization leads to a biased estimation of the evaluation metric (See Section~\ref{sec:exp:metric}).

\subsection{Evaluation metric}
\label{sec:exp:metric}

Intuitively, we prefer a saliency method that highlights features that align closely with the predictions (\eg highlights the object of interest in an image or the words indicating the sentiment of the sentence). To measure how well a saliency map achieves qualitative coherence, we use the fidelity metric \citep{dabkowski2017real}, defined as
\vspace{1pt}
\begin{equation}
SF(E(\cdot ; F^c), \bx) =-\log F^{c}(E(\bx ; F^c)\circ\bx )
 \label{eq:fidelitymetric}
\end{equation}
\vspace{1pt}
where $c$ indicates the predicted class of input $\bx$, and $E(\bx ; F^c)$ is the top-$K$-retained binary saliency map described above. $E(\bx ; F^c)\circ\bx$ performs entry-wise multiplication between $E(\bx ; F^c)$ and $\bx$, encoding the overlap between the object of interest and the concentration of the saliency map. The rationale behind this metric is that, by viewing the saliency score of a feature as its contribution to the predicted class, a good saliency method will highlight more important features and thus give rise to higher predicted class scores and lower metric values.

Note that, to guarantee a fair comparison among different saliency methods, it is important to retain the same number of important features for evaluation. Without such a scheme, pathologic cases such as $E(\bx ;F^c)=\mathbf{1}$ (\ie all saliency values equal to 1) would lead to highest fidelity score unexpectedly, which may be particularly problematic for alternative scheme such as 0-1 normalization.

\subsection{Performance in various applications}
\label{sec:exp:eval}

\subsubsection{Performance on the ImageNet dataset}
\label{sec:exp:imagenet}

To evaluate the effectiveness of \methodname{}, we first applied \methodname{} to randomly sampled images from the ImageNet dataset \citep{russakovsky2015imagenet}, with a pretrained VGG16 model \citep{simonyan2014very} (See Supplementary S7 for the applicability to diverse CNN architectures such as AlexNet \citep{krizhevsky2012imagenet} and ResNet \citep{he2016deep}). The $3 \times 3$ image patches are treated as swappable features in generating decoys. A side-by-side comparison (Figure~\ref{fig:imagenet}(A)) suggests that decoys consistently help to reduce noise and produce more visually coherent saliency maps. For example, the original integrated gradient method highlights the region of the dog's head in a scattered format, which is also revealed by the difference plot. In contrast, the decoy-enhanced integrated gradient method not only highlights the missing body but also identifies the dog's head with more details such as ears, cheek, and nose (See Supplementary Section S13 for more visualization examples). The visual coherence is also quantitatively supported by the saliency fidelity score.

To further evaluate the necessity of the two steps in our method (\ie decoy generation and aggregation), we carried out a control experiment by replacing each step with alternatives. Specifically, as alternatives to the decoy generation, we used an image in which all pixel values are either replaced with a single mean pixel value or contaminated with Gaussian white noise. For the decoy aggregation, we calculated the mean saliency score as the alternative. As shown in Figure~\ref{fig:imagenet}(B), our method, which incorporate both steps, yields the best performance. This validates the effectiveness of our two-step approach.

\begin{figure*}[th!]
\centering
\includegraphics[width=1.0\textwidth]{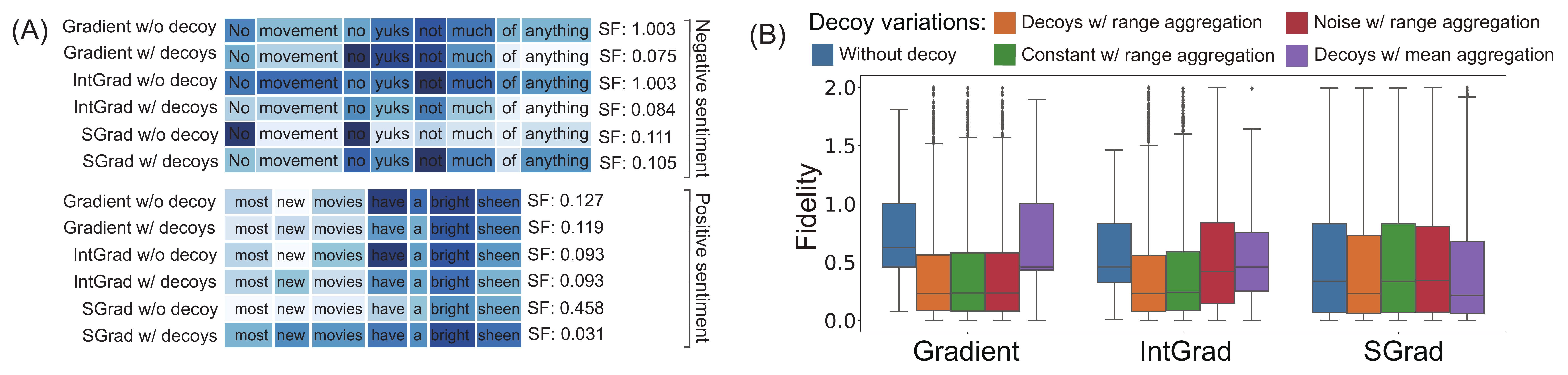}
\vspace{2pt}
\caption{{\bf Results obtained from the SST dataset.} (A) Visualization of saliency maps in each word, where the normalized saliency values are shown for better distinction. (B) Fidelity comparison of the original saliency method, our method, and its alternatives. Here, the alternative methods represent the practice of replacing the decoy generation (Equation~\eqref{eq:decoygen1}) with constant perturbation or noise perturbation as well as the practice of replacing the decoy aggregation (Equation~\eqref{eq:saliencystat}) with mean aggregation.  See Supplementary Section S14 for more statistics about the performance differences between our method and the baselines.}
\vspace{10pt}
\label{fig:sst}
\end{figure*}

Thirdly, to evaluate the computational  efficiency of \methodname{}, we carried out a fidelity comparison with respect to the number of decoys to optimize. As discussed in Section~\ref{sec:decoygen}, multiple swappable patches are aggregated into one combined patch mask for computational efficiency. Consequently, the mask multiplicity (\ie the number of swappable patches per mask) is inversely proportional to the number of decoys to optimize. Figure~\ref{fig:imagenet}(C) shows that our method achieves stable fidelity scores across a wide range of decoy numbers. Furthermore, as shown in Supplementary Section S10, the computational cost to optimize a single decoy in \methodname{} is negligible compared to even the fastest vanilla gradient-based saliency method. This analysis result confirms that our system could give reasonably good saliency maps without introducing too much computational cost.

Finally, In Supplementary Section S11, we run a sensitivity test on other hyper-parameters (\ie the swappable feature size $P$, the targeted network layer $\ell$, and the initial Lagrange multiplier $\lambda$). The results show that our method is insensitive to substantial variation of these hyperparameters. This is an important property because users do not need to extensively tune the hyper-parameters when using our method.


\subsubsection{Performance on the Stanford Sentiment Treebank (SST) dataset}
\label{sec:exp:sst}

To further evaluate the effectiveness of \methodname{}, we applied the method to randomly sampled sentences from the Stanford Sentiment Treebank (SST) \citep{russakovsky2015imagenet}. We trained a two-layer CNN \citep{kim2014convolutional} which takes the pretrained word embeddings as input~\citep{pennington2014glove} (See Supplementary Section S6 for more details about the experimental setup). As suggested by \citet{guan2019towards}, the average saliency value of all dimensions of a word embedding is regarded as the word-level saliency value. The embeddings of the words are treated as swappable features when generating decoys. As shown in Figure~\ref{fig:sst}(A), a side-by-side comparison suggests that our method consistently helps to produce semantically more meaningful saliency maps. For example, in a sentence with negative sentiment, keywords associated with negation, such as ``no" and ``not," are more strongly highlighted by decoy-enhanced saliency methods. The semantic coherence is also quantitatively supported by the saliency fidelity (Figure~\ref{fig:sst}(B)). We also tested the alternatives mentioned above: constant (replacing the decoy generation with the mean embedding of the whole dictionary) and noise perturbation with range aggregation, and decoys with mean aggregation. Figure~\ref{fig:sst}(B) shows that our method outperforms these alternatives.

To demonstrate the effectiveness of \methodname{} on models other than CNNs, we carried out experiments on a multi-layer perceptron trained with a network intrusion dataset. The results (Supplementary Section S8) are consistent with those on CNNs, thereby confirming our method's applicability to non-CNN architectures.
\begin{figure*}[th!]
\centering
\includegraphics[width=1.0\textwidth]{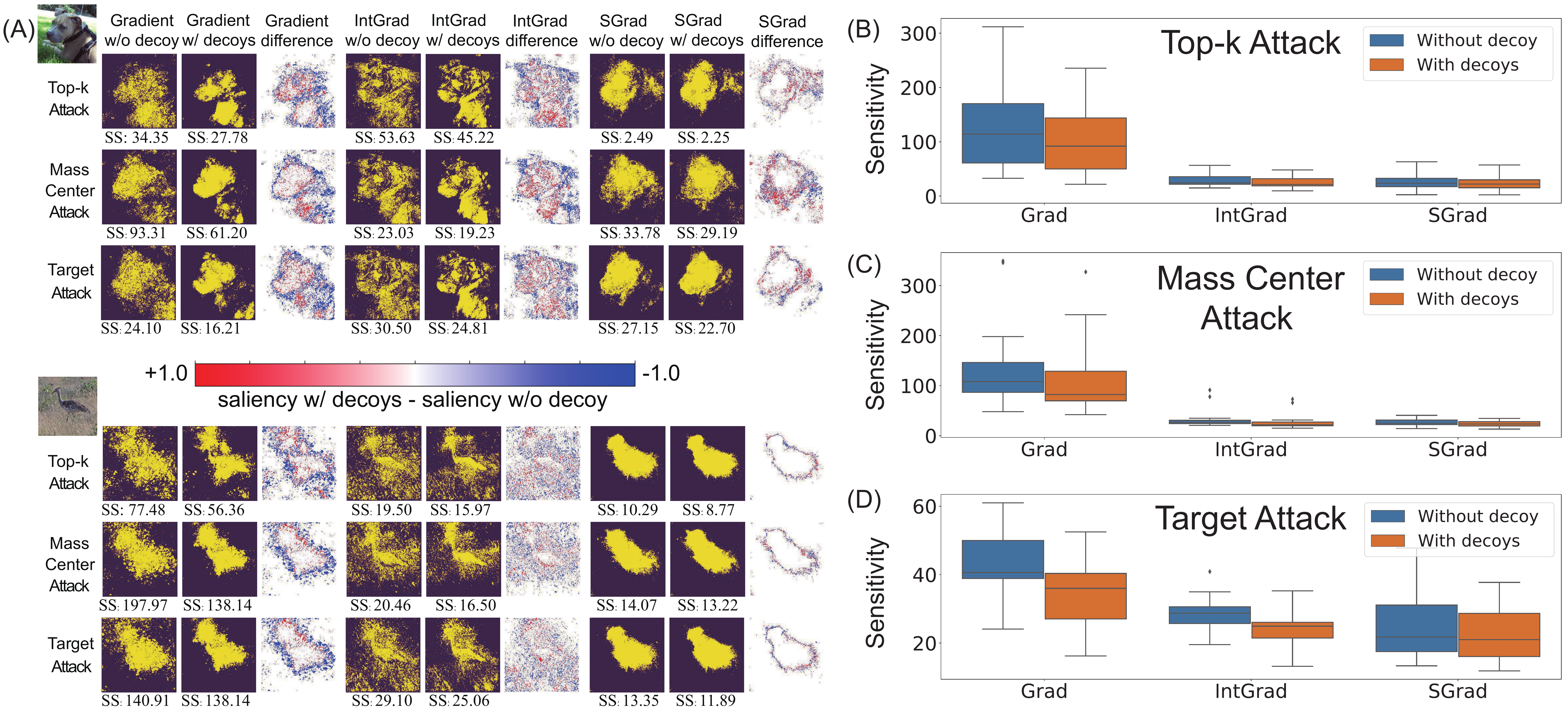}
\vspace{2pt}
\caption{{\bf Robustness to adversarial attacks on images.} (A) Visualization of saliency maps under adversarial attacks. (B)--(D) The decoy-enhanced saliency score is compared to the original saliency score under adversarial attacks, evaluated by sensitivity.  See Supplementary Section S14 for more statistics about the performance differences between our method and the baselines.}
\vspace{10pt}
\label{fig:sensitivity}
\end{figure*}
\subsection{Robustness to adversarial attacks}
\label{sec:exp:adversarial}

An important design philosophy of  \methodname{} is to model the variation of an input sample originating from either sensor noise or unknown perturbations by using decoys. We therefore hypothesized that \methodname{} may be particularly robust to adversarial manipulations of images. To test this hypothesis, we evaluated the robustness of our method to adversarial manipulations of images subject to three popular attacks \citep{ghorbani2017interpretation}:  (1) the top-$k$ attack, which seeks to decrease the scores of the top $k$ most important features, (2) the target attack, which aims to increase the importance of a pre-specified region in the input image, and (3) the mass-center attack, which aims to spatially change the mass center of the original saliency map. Here, we specify the bottom-right $4\times4$ region of the original image for the target attack and select $k=5000$ in the top-$k$ attack (See Supplementary Section S6 for detailed setups). We use the sensitivity metric \citep{alvarez2018towards} to quantify the robustness of a saliency method $E$ to attack, defined as:
\vspace{1pt}
\begin{equation}
SS(E(\cdot, F^c), \bx, \hat{\bx}) = \frac{\|(E(\bx, F^c) - E(\hat{\bx}, F^c))\|_2}{\|\bx - \hat{\bx}\|_2}
\end{equation}
\vspace{1pt}
where $\hat{\bx}$ is the perturbed image of $\bx$. A small $SS$ value means that similar inputs do not lead to substantially different saliency maps. As shown in Figure~\ref{fig:sensitivity}(A), a side-by-side comparison suggests that decoys consistently yield low sensitivity scores and help to produce more visually coherent saliency maps, mitigating the impact of various adversarial attacks (See the Supplementary material for more examples). The visual coherence and robustness to adversarial attacks are also quantitatively supported by Figure~\ref{fig:sensitivity}(B)--(D).

\vspace{10pt}
\section{Discussion and conclusion}

In this work, we propose \methodname{}, a method for computing, from a given saliency method, decoy-enhanced saliency scores that yield more accurate and robust saliency maps. We formulate the decoy generation as an optimization problem, applicable to diverse DNN architectures. We demonstrate the superior performance of our method relative to three standard saliency methods, both qualitatively and quantitatively, even in the presence of various adversarial perturbations to the image. From a theoretical perspective, by deriving a closed-form solution, we show that the proposed score can provably compensate for the limitations of existing saliency methods by reflecting the joint effects from other dependent features and maintaining robustness to adversarial perturbations.  We also demonstrate the computational efficiency of \methodname{}, and we show that the cost to optimize a single decoy is small, indicating that our technique can improve upon existing saliency methods without introducing too much computational overhead.

This work points to several promising directions for future research.
First, \methodname{} is designed for non-linear models such as feedforward DNNs which are most in need of interpretation. Future work will explore the extension of our method to other models (\eg linear model and recurrent neural networks) and to inputs with categorical or discrete features.
Second, recent work \citep{etmann2019connection,chen2019robust,chalasani2020concise} shows that adversarial training can improve a DNN's interpretability. It is worth exploring whether \methodname{} could further enhance the quality of saliency maps derived from these adversarially retrained classifiers.
Finally, a promising direction could be reframing interpretability as hypothesis testing and using decoys to deliver a set of salient features, subject to false discovery rate control at some pre-specified level \citep{burns2019hypothesis, lu2018deeppink}.	

\section*{Acknowledgments}
We would like to thank the anonymous reviewers and Meta reviewer for their helpful comments. This project was supported in part by NSF grant CNS-1718459, by NSF grant CNS-1954466, and by ONR grant N00014-20-1-2008.

\bibliographystyle{icml2021}
\bibliography{ref}

\begin{thebibliography}{54}
\providecommand{\natexlab}[1]{#1}
\providecommand{\url}[1]{\texttt{#1}}
\expandafter\ifx\csname urlstyle\endcsname\relax
  \providecommand{\doi}[1]{doi: #1}\else
  \providecommand{\doi}{doi: \begingroup \urlstyle{rm}\Url}\fi

\bibitem[Adebayo et~al.(2018)Adebayo, Gilmer, Muelly, Goodfellow, Hardt, and
  Kim]{adebayo2018sanity}
Adebayo, J., Gilmer, J., Muelly, M., Goodfellow, I., Hardt, M., and Kim, B.
\newblock Sanity checks for saliency maps.
\newblock In \emph{Proc. of NeurIPS}, 2018.

\bibitem[Alvarez-Melis \& Jaakkola(2018)Alvarez-Melis and
  Jaakkola]{alvarez2018towards}
Alvarez-Melis, D. and Jaakkola, T.~S.
\newblock Towards robust interpretability with self-explaining neural networks.
\newblock In \emph{Proc. of NeurIPS}, 2018.

\bibitem[Ancona et~al.(2018)Ancona, Ceolini, {\"O}ztireli, and
  Gross]{ancona2017towards}
Ancona, M., Ceolini, E., {\"O}ztireli, C., and Gross, M.
\newblock Towards better understanding of gradient-based attribution methods
  for deep neural networks.
\newblock In \emph{Proc. of ICLR}, 2018.

\bibitem[Barber \& Cand{\`e}s(2015)Barber and
  Cand{\`e}s]{barber2015controlling}
Barber, R.~F. and Cand{\`e}s, E.~J.
\newblock Controlling the false discovery rate via knockoffs.
\newblock \emph{The Annals of Statistics}, 2015.

\bibitem[Binder et~al.(2016)Binder, Montavon, Lapuschkin, M{\"u}ller, and
  Samek]{binder2016layer}
Binder, A., Montavon, G., Lapuschkin, S., M{\"u}ller, K.-R., and Samek, W.
\newblock Layer-wise relevance propagation for neural networks with local
  renormalization layers.
\newblock In \emph{Proc. of ICANN}, 2016.

\bibitem[Burns et~al.(2019)Burns, Thomason, and Tansey]{burns2019hypothesis}
Burns, C., Thomason, J., and Tansey, W.
\newblock Interpreting black box models via hypothesis testing.
\newblock \emph{arXiv:1904.00045}, 2019.

\bibitem[Chalasani et~al.(2020)Chalasani, Chen, Chowdhury, Wu, and
  Jha]{chalasani2020concise}
Chalasani, P., Chen, J., Chowdhury, A.~R., Wu, X., and Jha, S.
\newblock Concise explanations of neural networks using adversarial training.
\newblock In \emph{Proc. of ICML}, 2020.

\bibitem[Chang et~al.(2019)Chang, Creager, Goldenberg, and
  Duvenaud]{change2019xplaining}
Chang, C.-H., Creager, E., Goldenberg, A., and Duvenaud, D.
\newblock Explaining image classifiers by counterfactual generation.
\newblock In \emph{Proc. of ICLR}, 2019.

\bibitem[Chen et~al.(2019{\natexlab{a}})Chen, Li, Tao, Barnett, Rudin, and
  Su]{chen2019looks}
Chen, C., Li, O., Tao, D., Barnett, A., Rudin, C., and Su, J.~K.
\newblock This looks like that: deep learning for interpretable image
  recognition.
\newblock In \emph{Proc. of NeurIPS}, 2019{\natexlab{a}}.

\bibitem[Chen et~al.(2018)Chen, Song, Wainwright, and Jordan]{chen2018learning}
Chen, J., Song, L., Wainwright, M.~J., and Jordan, M.~I.
\newblock Learning to explain: An information-theoretic perspective on model
  interpretation.
\newblock In \emph{Proc. of ICML}, 2018.

\bibitem[Chen et~al.(2019{\natexlab{b}})Chen, Song, Wainwright, and
  Jordan]{chen2018shapley}
Chen, J., Song, L., Wainwright, M.~J., and Jordan, M.~I.
\newblock L-shapley and c-shapley: Efficient model interpretation for
  structured data.
\newblock In \emph{Proc. of ICLR}, 2019{\natexlab{b}}.

\bibitem[Chen et~al.(2019{\natexlab{c}})Chen, Wu, Rastogi, Liang, and
  Jha]{chen2019robust}
Chen, J., Wu, X., Rastogi, V., Liang, Y., and Jha, S.
\newblock Robust attribution regularization.
\newblock In \emph{Proc. of NeurIPS}, 2019{\natexlab{c}}.

\bibitem[Dabkowski \& Gal(2017)Dabkowski and Gal]{dabkowski2017real}
Dabkowski, P. and Gal, Y.
\newblock Real time image saliency for black box classifiers.
\newblock In \emph{Proc. of NeurIPS}, 2017.

\bibitem[Etmann et~al.(2019)Etmann, Lunz, Maass, and
  Sch{\"o}nlieb]{etmann2019connection}
Etmann, C., Lunz, S., Maass, P., and Sch{\"o}nlieb, C.-B.
\newblock On the connection between adversarial robustness and saliency map
  interpretability.
\newblock \emph{arXiv preprint arXiv:1905.04172}, 2019.

\bibitem[Fan et~al.(2017)Fan, Zhao, and Ermon]{fan2017adversarial}
Fan, L., Zhao, S., and Ermon, S.
\newblock Adversarial localization network.
\newblock In \emph{Proc. of NeurIPS LLD Workshop}, 2017.

\bibitem[Fong \& Vedaldi(2017)Fong and Vedaldi]{fong2017interpretable}
Fong, R.~C. and Vedaldi, A.
\newblock Interpretable explanations of black boxes by meaningful perturbation.
\newblock In \emph{Proc. of ICCV}, 2017.

\bibitem[Ghorbani et~al.(2017)Ghorbani, Abid, and
  Zou]{ghorbani2017interpretation}
Ghorbani, A., Abid, A., and Zou, J.
\newblock Interpretation of neural networks is fragile.
\newblock \emph{arXiv:1710.10547}, 2017.

\bibitem[Ghorbani et~al.(2019)Ghorbani, Wexler, Zou, and
  Kim]{ghorbani2019towards}
Ghorbani, A., Wexler, J., Zou, J.~Y., and Kim, B.
\newblock Towards automatic concept-based explanations.
\newblock In \emph{Proc. of NeurIPS}, 2019.

\bibitem[Goyal et~al.(2019)Goyal, Wu, Ernst, Batra, Parikh, and
  Lee]{goyal2019counterfactual}
Goyal, Y., Wu, Z., Ernst, J., Batra, D., Parikh, D., and Lee, S.
\newblock Counterfactual visual explanations.
\newblock \emph{Proc. of ICML}, 2019.

\bibitem[Guan et~al.(2019)Guan, Wang, Zhang, Chen, He, and
  Xie]{guan2019towards}
Guan, C., Wang, X., Zhang, Q., Chen, R., He, D., and Xie, X.
\newblock Towards a deep and unified understanding of deep neural models in
  nlp.
\newblock In \emph{Proc. of ICML}, 2019.

\bibitem[He et~al.(2016)He, Zhang, Ren, and Sun]{he2016deep}
He, K., Zhang, X., Ren, S., and Sun, J.
\newblock Deep residual learning for image recognition.
\newblock In \emph{Proc. of CVPR}, 2016.

\bibitem[Hendrycks \& Dietterich(2019)Hendrycks and
  Dietterich]{hendrycks2019benchmarking}
Hendrycks, D. and Dietterich, T.
\newblock Benchmarking neural network robustness to common corruptions and
  perturbations.
\newblock In \emph{Proc. of ICLR}, 2019.

\bibitem[Hooker et~al.(2019)Hooker, Erhan, Kindermans, and
  Kim]{hooker2019benchmark}
Hooker, S., Erhan, D., Kindermans, P.-J., and Kim, B.
\newblock A benchmark for interpretability methods in deep neural networks.
\newblock In \emph{Proc. of NeurIPS}, 2019.

\bibitem[Ikeno \& Hara(2018)Ikeno and Hara]{ikeno2018maximizing}
Ikeno, K. and Hara, S.
\newblock Maximizing invariant data perturbation with stochastic optimization.
\newblock \emph{arXiv preprint arXiv:1807.05077}, 2018.

\bibitem[Kim(2014)]{kim2014convolutional}
Kim, Y.
\newblock Convolutional neural networks for sentence classification.
\newblock \emph{Proc. of EMNLP}, 2014.

\bibitem[Kindermans et~al.(2017)Kindermans, Hooker, Adebayo, Alber, Sch{\"u}tt,
  D{\"a}hne, Erhan, and Kim]{kindermans2017reliability}
Kindermans, P.-J., Hooker, S., Adebayo, J., Alber, M., Sch{\"u}tt, K.~T.,
  D{\"a}hne, S., Erhan, D., and Kim, B.
\newblock The ({Un}) reliability of saliency methods.
\newblock \emph{arXiv:1711.00867}, 2017.

\bibitem[Koh \& Liang(2017)Koh and Liang]{koh2017understanding}
Koh, P.~W. and Liang, P.
\newblock Understanding black-box predictions via influence functions.
\newblock \emph{Proc. of ICML}, 2017.

\bibitem[Koh et~al.(2019)Koh, Ang, Teo, and Liang]{koh2019accuracy}
Koh, P. W.~W., Ang, K.-S., Teo, H., and Liang, P.~S.
\newblock On the accuracy of influence functions for measuring group effects.
\newblock In \emph{Proc. of NeurIPS}, 2019.

\bibitem[Krizhevsky et~al.(2012)Krizhevsky, Sutskever, and
  Hinton]{krizhevsky2012imagenet}
Krizhevsky, A., Sutskever, I., and Hinton, G.~E.
\newblock Imagenet classification with deep convolutional neural networks.
\newblock In \emph{Proc. of NeurIPS}, 2012.

\bibitem[Levine et~al.(2019)Levine, Singla, and Feizi]{levine2019certifiably}
Levine, A., Singla, S., and Feizi, S.
\newblock Certifiably robust interpretation in deep learning.
\newblock \emph{arXiv preprint arXiv:1905.12105}, 2019.

\bibitem[Lipton(2016)]{lipton2016mythos}
Lipton, Z.~C.
\newblock The mythos of model interpretability.
\newblock \emph{arXiv:1606.03490}, 2016.

\bibitem[Lu et~al.(2018)Lu, Fan, Lv, and Noble]{lu2018deeppink}
Lu, Y., Fan, Y., Lv, J., and Noble, W.~S.
\newblock {DeepPINK}: reproducible feature selection in deep neural networks.
\newblock In \emph{Proc. of NeurIPS}, 2018.

\bibitem[Lundberg \& Lee(2017)Lundberg and Lee]{lundberg2017unified}
Lundberg, S.~M. and Lee, S.-I.
\newblock A unified approach to interpreting model predictions.
\newblock In \emph{Proc. of NeurIPS}, 2017.

\bibitem[Moosavi-Dezfooli et~al.(2017)Moosavi-Dezfooli, Fawzi, Fawzi, and
  Frossard]{moosavi2017universal}
Moosavi-Dezfooli, S.-M., Fawzi, A., Fawzi, O., and Frossard, P.
\newblock Universal adversarial perturbations.
\newblock In \emph{Proc. of CVPR}, 2017.

\bibitem[Nie et~al.(2018)Nie, Zhang, and Patel]{nie2018theoretical}
Nie, W., Zhang, Y., and Patel, A.
\newblock A theoretical explanation for perplexing behaviors of
  backpropagation-based visualizations.
\newblock In \emph{Proc. of ICML}, 2018.

\bibitem[Pennington et~al.(2014)Pennington, Socher, and
  Manning]{pennington2014glove}
Pennington, J., Socher, R., and Manning, C.~D.
\newblock Glove: Global vectors for word representation.
\newblock In \emph{Proc. of EMNLP}, 2014.

\bibitem[Ribeiro et~al.(2016)Ribeiro, Singh, and Guestrin]{ribeiro2016should}
Ribeiro, M.~T., Singh, S., and Guestrin, C.
\newblock Why should i trust you?: Explaining the predictions of any
  classifier.
\newblock In \emph{Proc. of KDD}, 2016.

\bibitem[Russakovsky et~al.(2015)Russakovsky, Deng, Su, Krause, Satheesh, Ma,
  Huang, Karpathy, Khosla, Bernstein, et~al.]{russakovsky2015imagenet}
Russakovsky, O., Deng, J., Su, H., Krause, J., Satheesh, S., Ma, S., Huang, Z.,
  Karpathy, A., Khosla, A., Bernstein, M., et~al.
\newblock Imagenet large scale visual recognition challenge.
\newblock \emph{International Journal of Computer Vision}, 2015.

\bibitem[Schulz et~al.(2020)Schulz, Sixt, Tombari, and
  Landgraf]{schulz2020restricting}
Schulz, K., Sixt, L., Tombari, F., and Landgraf, T.
\newblock Restricting the flow: Information bottlenecks for attribution.
\newblock 2020.

\bibitem[Selvaraju et~al.(2016)Selvaraju, Das, Vedantam, Cogswell, Parikh, and
  Batra]{selvaraju2016grad}
Selvaraju, R.~R., Das, A., Vedantam, R., Cogswell, M., Parikh, D., and Batra,
  D.
\newblock Grad-cam: Visual explanations from deep networks via gradient-based
  localization.
\newblock \emph{arXiv:1611.07450}, 2016.

\bibitem[Shrikumar et~al.(2017)Shrikumar, Greenside, and
  Kundaje]{shrikumar2017learning}
Shrikumar, A., Greenside, P., and Kundaje, A.
\newblock Learning important features through propagating activation
  differences.
\newblock In \emph{Proc. of ICML}, 2017.

\bibitem[Simonyan \& Zisserman(2014)Simonyan and Zisserman]{simonyan2014very}
Simonyan, K. and Zisserman, A.
\newblock Very deep convolutional networks for large-scale image recognition.
\newblock \emph{arXiv:1409.1556}, 2014.

\bibitem[Simonyan et~al.(2013)Simonyan, Vedaldi, and
  Zisserman]{simonyan2013deep}
Simonyan, K., Vedaldi, A., and Zisserman, A.
\newblock Deep inside convolutional networks: Visualising image classification
  models and saliency maps.
\newblock \emph{arXiv:1312.6034}, 2013.

\bibitem[Singla et~al.(2019)Singla, Wallace, Feng, and
  Feizi]{singla2019understanding}
Singla, S., Wallace, E., Feng, S., and Feizi, S.
\newblock Understanding impacts of high-order loss approximations and features
  in deep learning interpretation.
\newblock \emph{arXiv:1902.00407}, 2019.

\bibitem[Sixt et~al.(2020)Sixt, Granz, and Landgraf]{sixt2019explanations}
Sixt, L., Granz, M., and Landgraf, T.
\newblock When explanations lie: Why many modified bp attributions fail.
\newblock In \emph{Proc. of ICML}, 2020.

\bibitem[Smilkov et~al.(2017)Smilkov, Thorat, Kim, Vi{\'e}gas, and
  Wattenberg]{smilkov2017smoothgrad}
Smilkov, D., Thorat, N., Kim, B., Vi{\'e}gas, F., and Wattenberg, M.
\newblock Smoothgrad: removing noise by adding noise.
\newblock \emph{arXiv:1706.03825}, 2017.

\bibitem[Springenberg et~al.(2014)Springenberg, Dosovitskiy, Brox, and
  Riedmiller]{springenberg2014striving}
Springenberg, J.~T., Dosovitskiy, A., Brox, T., and Riedmiller, M.
\newblock Striving for simplicity: The all convolutional net.
\newblock \emph{arXiv preprint arXiv:1412.6806}, 2014.

\bibitem[Sturmfels et~al.(2020)Sturmfels, Lundberg, and
  Lee]{sturmfels2020visualizing}
Sturmfels, P., Lundberg, S., and Lee, S.-I.
\newblock Visualizing the impact of feature attribution baselines.
\newblock \emph{Distill}, 2020.

\bibitem[Sundararajan et~al.(2017)Sundararajan, Taly, and
  Yan]{sundararajan2017axiomatic}
Sundararajan, M., Taly, A., and Yan, Q.
\newblock Axiomatic attribution for deep networks.
\newblock In \emph{Proc. of ICML}, 2017.

\bibitem[Toneva \& Wehbe(2019)Toneva and Wehbe]{toneva2019interpreting}
Toneva, M. and Wehbe, L.
\newblock Interpreting and improving natural-language processing (in machines)
  with natural language-processing (in the brain).
\newblock In \emph{Proc. of NeurIPS}, 2019.

\bibitem[Yeh et~al.(2018)Yeh, Kim, Yen, and Ravikumar]{yeh2018representer}
Yeh, C.-K., Kim, J., Yen, I. E.-H., and Ravikumar, P.~K.
\newblock Representer point selection for explaining deep neural networks.
\newblock In \emph{Proc. of NeurIPS}, 2018.

\bibitem[Yousefzadeh \& O'Leary(2019)Yousefzadeh and
  O'Leary]{yousefzadeh2019interpreting}
Yousefzadeh, R. and O'Leary, D.~P.
\newblock Interpreting neural networks using flip points.
\newblock \emph{arXiv preprint arXiv:1903.08789}, 2019.

\bibitem[Zeiler \& Fergus(2014)Zeiler and Fergus]{zeiler2014visualizing}
Zeiler, M.~D. and Fergus, R.
\newblock Visualizing and understanding convolutional networks.
\newblock In \emph{Proc. of ECCV}, 2014.

\bibitem[Zo{\l}na et~al.(2019)Zo{\l}na, Geras, and Cho]{zolna2019classifier}
Zo{\l}na, K., Geras, K.~J., and Cho, K.
\newblock Classifier-agnostic saliency map extraction.
\newblock In \emph{Proceedings of AAAI}, 2019.

\end{thebibliography}

\end{document}